\documentclass[a4paper]{article}
\usepackage[utf8]{inputenc}
\usepackage{authblk}
\usepackage{placeins}
\usepackage{graphicx}
\usepackage{subcaption}
\usepackage{amssymb}
\usepackage{amsmath}
\usepackage{makecell}
\usepackage[dvipsnames]{xcolor}
\usepackage[T1]{fontenc}
\usepackage{xspace}
\usepackage{printlen}

\usepackage{hyperref}
\hypersetup{
    colorlinks=true,
    linkcolor=blue,
    filecolor=magenta,      
    urlcolor=cyan,
    citecolor=Green
}

\newcommand{\etal}{\textit{et al}. }

\newcommand{\PresQ}[0]{\textsc{PresQ}\xspace}

\title{
    Two-sample test based on Self-Organizing Maps
    \thanks{This research was funded by the Spanish AEI (DOI:10.13039/501100011033)
        through the project CR\^EPES (ref. PID2020-115844RB-I00) with ERDF funds.}
}

\author[1]{
    Alejandro Álvarez-Ayllón
    \thanks{
        Corresponding author:
        \texttt{\href{mailto:alejandro.alvarez@uca.es}{alejandro.alvarez@uca.es}}
    }
}
\author[1]{Manuel Palomo-Duarte}
\author[1]{Juan-Manuel Dodero}

\affil[1]{Department of Computer Science and Engineering, University of Cadiz}

\date{}

\uselengthunit{in}

\begin{document}

\maketitle

\begin{abstract}
    Machine-learning classifiers can be leveraged as a two-sample statistical test.
    Suppose each sample is assigned a different label and that a classifier can obtain
    a better-than-chance result discriminating them. In this case, we can infer that
    both samples originate from different populations.
    However, many types of models, such as neural networks, behave as a black-box
    for the user: they can reject that both samples originate from the same population,
    but they do not offer insight into how both samples differ.
    Self-Organizing Maps are a dimensionality reduction initially devised as a
    data visualization tool that displays emergent properties, being also useful for
    classification tasks. Since they can be used as classifiers, they can be used also
    as a two-sample statistical test. But since their original purpose is visualization,
    they can also offer insights.
\end{abstract}

\section{Introduction}

A classification task can be seen as a sort of two-sample statistical test. If a classifier
trained over two samples can distinguish between them with better-than-chance performance,
effectively the classifier rejects the null hypothesis that both samples come from the
same distribution~\cite{friedman2004multivariate}.

While a formal statistical test can be more suitable when the alternative
hypothesis is known, machine learning classifiers can be a good alternative
when the data is complex and
abundant~\cite{kirchler2020two,kim2021classification,pmlr-v119-liu20m}.

Furthermore, the representation ``learned'' by some classifiers can be helpful to
examine how the samples differ~\cite{friedman2004multivariate,lopez2016revisiting}, or it
could be used later for other purposes, such as directly classifying future samples.

\medskip

A Self-Organizing Map (SOM) is a technique for dimensionality
reduction. It is a type of neural network that learns a low-dimension representation
- generally 2D - of the original high dimensional space while maintaining the topological
layout of the original data~\cite{kohonen1982self,Villmann1999}.

The learned map can be directly used for unsupervised clustering - when the map is big enough -
and classification tasks if the training data is labeled~\cite{ultsch2005esom,ultsch2007emergence}.
For instance, the neurons can be labeled according to the training data labels mapped into
their region. During classification, objects can be labeled according to the label of the neuron
into which they are mapped.
Thus, SOMs can be used as a building block of an ML-based two-sample test, similar to
\emph{k-Nearest Neighbors} or \emph{Neural Network} classifiers, with the valuable addition
of producing a representation that can be visualized.

\paragraph{Contribution:} In this paper, we propose a multivariate statistical test based on
Self-Organizing Maps that shows performance comparable to other techniques based on 
machine learning models and even superior for medium to big sample sizes. In addition to the 
$p$ value for $H_0: P = Q$, the test also outputs a trained SOM model that can be of
use for other tasks, such as classification or visualization.
Our proposal uses a $\chi^2$ statistic to compare the densities of both samples on
the projected plane instead of relying on a training-testing split. This allows us to fully
utilize the sample data. To our knowledge, using Self-Organizing Maps
to perform multidimensional two-sample testing has not been proposed
before~\cite{kaski1998bibliography,oja_bibliography_2003,polla_bibliography_2006}.

\paragraph{Motivating example} \PresQ is an algorithm that automatically searches
for subsets of attributes equally distributed between two datasets~\cite{alvarez2021noise}.
It relies on statistical tests. A test based on SOM provides, in addition to a $p$ value,
a trained projection that we can later use for binning and cross-matching both datasets using
the matching set of features.

\paragraph{Paper organization:} First, in section~\ref{sec:chi2}, we describe our proposal
for a multidimensional non-parametric statistical test based on Self-Organizing Maps.
In section~\ref{sec:background}, we introduce previous proposals based on machine-learning
techniques for non-parametric tests, together with their motivations.
In section~\ref{sec:exp_setup}, we describe the experimental setup we have used
to validate our proposal --- including the parametrization of the existing techniques
evaluated as a baseline ---. In section~\ref{sec:results} we present the results.
Later, in section~\ref{sec:discussion}, we discuss our interpretation of the results.
Finally, in section~\ref{sec:conclusions}, we compile our conclusions and propose areas for future
work.

\section{Background}
\label{sec:background}

\subsection{Classifier two-sample tests}
\label{sec:classifier2sample}
A binary classifier can be seen as a two-sample test. If a classifier has a better-than-chance
performance, it can be inferred that the two classes do not originate from the same underlying
population~\cite{friedman2004multivariate}. 

More formally, given two sets, $X = \{x_0,x_1,\ldots,x_n\}$ sampled from $P$, and \linebreak
$Z = \{z_1,z_2,\ldots,z_m\}$ sampled from $Q$. A test statistic $\hat t \sim T$ is
used to ``summarize'' the difference between both samples and, depending on a pre-established
significance level $\alpha$, used to reject the null hypothesis $H_0: P = Q$ if
$\alpha > P(T \ge \hat t | H_0)$.

When using a binary classifier for performing a statistical test, both samples
are pooled together $U = \{u_i\}_{i=1}^{n+m} = \{x_i\}_{i=1}^n \cup \{z_i\}_{i=1}^m$.
The samples originating from $P$ are labeled $y_i=1$ and the samples originating from $Q$, $y_i=-1$.

The original proposal trains a classifier on the \emph{complete} pooled sample.
This classifier is then used to score each data point, generating a set of scores
for the first sample $S_+$, and for the second $S_-$. The multi-dimensional comparison is thus
reduced to a regular univariate two-sample test problem~\cite{friedman2004multivariate}.

Another approach is to split the pooled dataset $\{u_i\}_{i=1}^{n+m}$ into training and
testing sets. A classifier is then trained on the former, and the accuracy is measured
for the latter. The accuracy becomes the test statistic $\hat t$ ,
which follows asymptotically $N(\frac{1}{2}, \frac{1}{4 n_{test}})$~\cite{lopez2016revisiting}. Alternatively, a permutation test can be used instead~\cite{kim2021classification}.
Two disadvantages of this kind of test are that they can not use the full sample for computing
the test statistic --- therefore, they are not suitable for small data sets ---
and they are underpowered due to the discrete nature of the test statistic~\cite{rosenblatt2021better}.

\subsection{Self-Organizing Map}

Self-Organizing Map (SOM) is an unsupervised machine-learning algorithm that \emph{learns}
a projection from a high-dimension input space into a low-dimension output space,
generally two-dimensional, to aid visualization.
The output space is modeled as a grid of \emph{neurons} --- a neural map --- that
\emph{respond} to a set of values from the input space~\cite{kohonen1982self}.
The output model preserves the topology of the input space: any continuous changes in the
input data cause a continuous change on the neural map~\cite{Villmann1999}.
In other words, input values close in the original high-dimensional space
trigger \emph{neurons} that are close in the low-dimensional projection~\cite{KOHONEN201352}.

Generally, the output space $W$ has to be defined before the training phase. The user needs to
define the shape of the grid --- square or hexagonal ---, its size, and whether
the map \emph{wraps around} (toroidal maps).
Each neuron $i$ from the model has an associated weight vector with the same dimensionality
as the input space, $w_{i}(t)$, where $t$ corresponds to the \emph{epoch} of the training stage.
The initial values $w_i(t_0)$ can be assigned randomly, or based on
Principal Component Analysis~\cite{KOHONEN201352}.

During the training, at each epoch $t$, each point $x$ from the training set --- or from a batch ---
is mapped to its \emph{best matching unit} (BMU), which is just the neuron whose weight vector is
the closest given a distance metric $d$:

\begin{equation}
    \operatorname{bmu}(x) = \underset{w_i \in W}{\operatorname{argmin}} \; d(x, w_i)
\end{equation}

Once this is done, the BMU and the weight of the neighboring neurons are updated so they become
\emph{closer} to the input data point:

\begin{equation}
    w_i(t + 1) = w_i(t) + \alpha h_{i,b}(t) (x - w_b(t))
\end{equation}

Where $0 \le \alpha \le 1$ is a learning factor that may or may not depend on $t$,
and $0 \le h_{i,b} \le 1$ is the neighborhood function, with usually a Gaussian shape
that shrinks at each epoch~\cite{Villmann1999,wittek2013somoclu}.

\begin{equation}
    h_{i,b}(t) = \exp(- \frac{||w_i - w_b||}{\delta(t)})
\end{equation}

This process can be repeated for multiple epochs or until convergence.

\medskip

Thanks to the preservation of topology, when the SOM grid is large enough, they display
emergent properties: they can be directly used for clustering, classification, and other
machine learning techniques. These are called Emergent Self-Organizing Maps (ESOM)~\cite{ultsch2005esom}. This motivates our proposal since if a statistical test
based on a SOM projection rejects the null hypothesis that two samples are equally distributed,
unlike other methods, it can provide insights into how they differ.

\section{\texorpdfstring{$\chi^2$}{χ²} test on the projection of the samples over a Self-Organizing Map}
\label{sec:chi2}

Thanks to the topology preservation of Self-Organizing maps, a classifier can be trained
on the output space rather than the input space. For instance, for a $k$-nearest neighbor ($k$NN)
approach, neurons can be labeled using the training data and a majority rule. Later, test data
can be assigned the label from its BMU. This is almost equivalent to a $k$NN classifier with $k=1$.
Furthermore, neurons belonging to sparse regions can be left unlabeled, so test data projected
into them can be labeled as \emph{unknown class}~\cite{ultsch2005esom,silva2011som}.

While a SOM-based classifier could be used in place of the neural or $k$NN classifiers proposed
originally~\cite{lopez2016revisiting}, we propose a different approach that does not require
splitting the input data into training and testing sets and leverages the distribution of the
data on the output space instead. The intuition behind this is that if two samples are equally
distributed on the original space, they must be equally distributed on the output space.

More specifically, our method works as follows:

\begin{enumerate}
    \item We train a Self-Organizing map $M$ of size $(w, h)$ over $U = X \cup Z$
    \item We project $X$ and $Z$ separately over the SOM $M$
    \item We compute how many points from $X$ and how many from $Z$ are mapped to a given neuron $n_i$
\end{enumerate}

\begin{align}
    R_i = \sum_{x \in X} [ \operatorname{bmu}(x) = i ] && S_i = \sum_{z \in Z} [ \operatorname{bmu}(z) = i ]
\end{align}

\begin{enumerate}
    \setcounter{enumi}{3}
    \item Finally, we perform a a $\chi^2$ two sample test comparing the counts for both samples
    on the output space
\end{enumerate}

\begin{equation}
    \label{eq:chi2}
    \chi^2 = \sum_{i=1}^{w \times h}{ \left\{ \frac{(K_1 R_i - K_2 S_i)^2}{R_i + S_i} [ R_i + S_i > 0 ] \right\}}
\end{equation}

Where $K_1$ and $K_2$ are two constants used to adjust for different sample sizes:

\begin{align}
    \label{eq:k1k2} K_1 = \sqrt{\frac{|Y|}{|X|}} && K_2 = \sqrt{\frac{|X|}{|Y|}}
\end{align}

Note that we ignore the bins where there are 0 objects. Under the null hypothesis
(both histograms are equal) the test statistic $\chi^2$ follows a $\chi^2$ distribution
with $k - c$ degrees of freedom,
where $k$ is the number of cells where ${R_i + S_i > 0}$, and $c = 1$ if the sample sizes are equal,
or $c = 0$ otherwise~\cite{press1993numerical}.

As with any test based on binning, its main disadvantages are that its results may
depend on the binning (in this case, the size of the SOM), and it requires more data points.

On the other hand, since the SOM \textit{adapts} to the topology of the original data,
it is less susceptible to artifacts than a simple 2D histogram due to the binning.

In any case, and very much like the motivation behind the classifier tests, as a side effect
of the test, we are left with a trained model that can be used for (1) visualization; and (2) for
big enough SOM and samples, even for clustering \cite{ultsch2005esom}.

Unlike other classifier tests, with our proposal, the whole dataset can be used for computing the
statistic~\cite{kirchler2020two}. Additionally, thanks to the regularization terms shown
in equation \ref{eq:k1k2}, it also works with unbalanced sample sizes, an
advantage over most kernel-based methods~\cite{song2021fast}.

We have implemented our proposal using Somuclu, a parallel tool for training self-organizing maps
on large data sets~\cite{wittek2013somoclu}.

The following section will describe the experimental setup used to evaluate our proposal.
Later, in section~\ref{sec:results} we will report the results of our tests.

\section{Experimental setup}
\label{sec:exp_setup}

\subsection{Evaluated alternatives}

We have considered four different two-sample tests based on machine learning techniques.
All of them have in common the merging of both samples into a single set $Z$,
labeled with 1 if the sample comes from $X$ or -1 if comes from $Y$.

\paragraph{Nearest neighbor type coincidences}

The assumption under $H_0$ is that on the neighboring area of any point, the number of samples
belonging to $X$ and to $Y$ should be similar, while if $f \neq g$, then there will be areas with
a higher density of objects coming from one of the two sets~\cite{Henze1988,Schilling1986b}.

To perform the test, consider a neighbor $r$ of a sample $z_i \in Z$. We set

\begin{equation}
\begin{split}
    I_i(r) &= 1, \textrm{ if the label of } z_i \textrm{ and of } z_r \textrm{ match }\\
    I_i(r) &= 0, \textrm{otherwise}
\end{split}
\end{equation}

The statistical test:

\begin{equation}
    T_{n,k} = \sum_{i=0}^{n}\sum_{r=0}^{k} I_i(r)
\end{equation}

Where $n$ is the total number of samples, and $k$ is the number of neighbors considered.
The distribution of the statistic is empirically obtained by applying a permutation test.
 
\paragraph{Classifier two-sample tests}
We have implemented the classifier two-sample test as described in section~\ref{sec:classifier2sample}
using \texttt{scikit-learn}~\cite{scikit-learn} neural
classifier\footnote{\texttt{sklearn.neural\_network.MLPClassifier}},
and $k$NN classifier\footnote{\texttt{sklearn.neighbors.KNeighborsClassifier}}
with their default parameterization. Table~\ref{tab:classifier_diff} summarizes the differences
with respect to the original proposal. However, these differences should not affect significantly
the performance~\cite{lopez2016revisiting}.

\begin{table}[htpb]
\centering
\begin{tabular}{lrr}
\multicolumn{1}{c}{\bfseries Parameter}       & \bfseries Revisiting\ldots & \texttt{scikit-learn}     \\ \hline
\multicolumn{3}{c}{\bfseries C2ST-NN} \\
Number of hidden layers &  1     &   1    \\
Number of neurons       & 20     & 100    \\
Activation              & ReLU   & ReLU   \\
Optimizer               & Adam   & Adam   \\
Epochs                  & 100    & 200    \\
\multicolumn{3}{c}{\bfseries C2ST-kNN} \\
$k$                     & $|X|/2$ & 5 \\
\end{tabular}
\caption{Differences between our parameterization (\texttt{scikit-learn} defaults) and the one
used in the original proposal~\cite{lopez2016revisiting}.}
\label{tab:classifier_diff}
\end{table}

\paragraph{Kernel Methods}
Kernel two-sample tests are based on computing the maximum mean discrepancy (MMD) between the samples,
which is the distance between their expected features in a reproducing kernel Hilbert space (RKHS).
The original proposal~\cite{gretton2012kernel}, however, is computationally
expensive to compute the test statistic --- $O(N^2)$ --- and to approximate its distribution under
$H_0$ --- $O(N^2)$ or $O(N^3)$ depending on the method~\cite{zaremba2013b}.
A proposed alternative, MMD-B~\cite{zaremba2013b}, splits the input data into blocks, computes the original,
unbiased MMD statistic on each block --- which are i.i.d ---, and averages the results.
Because of the central limit theorem, this average follows asymptotically a normal distribution.
Song \etal~\cite{song2021fast} propose another test statistic based on MMD-B that allows
unbalanced sample sizes and is more robust to the chosen kernel bandwidth (i.e., $\sigma$ on a Gaussian kernel).

\section{Results}
\label{sec:results}

We have performed five experiments to evaluate the performance of our SOM
two-sample test proposal.

For the first three setups - Normal, DC2, and the \emph{Open University Learning Analytics} dataset
- we set the significance level $\alpha = 0.1$. We then measure the run-time and empirical 
type I and type II error rates over $200$ repeated tests for all the evaluated
tests: (1) SOM (our proposal), (2) the kNN permutation test~\cite{Schilling1986b},
and (3) two classifier tests (kNN and Neural Network)~\cite{lopez2016revisiting}.

These values are measured for: (1) a fixed sample size of $n = m = 500$ and variable dimension;
and (2) for variable sample sizes and full dimensionality.

We have used a \emph{K-Best} feature selection to decide in which order dimensions are added.
Therefore, increasing the dimensionality is expected to have a diminishing return.

\subsection{Normal}
\begin{figure}[htpb]
    \centering
    \includegraphics{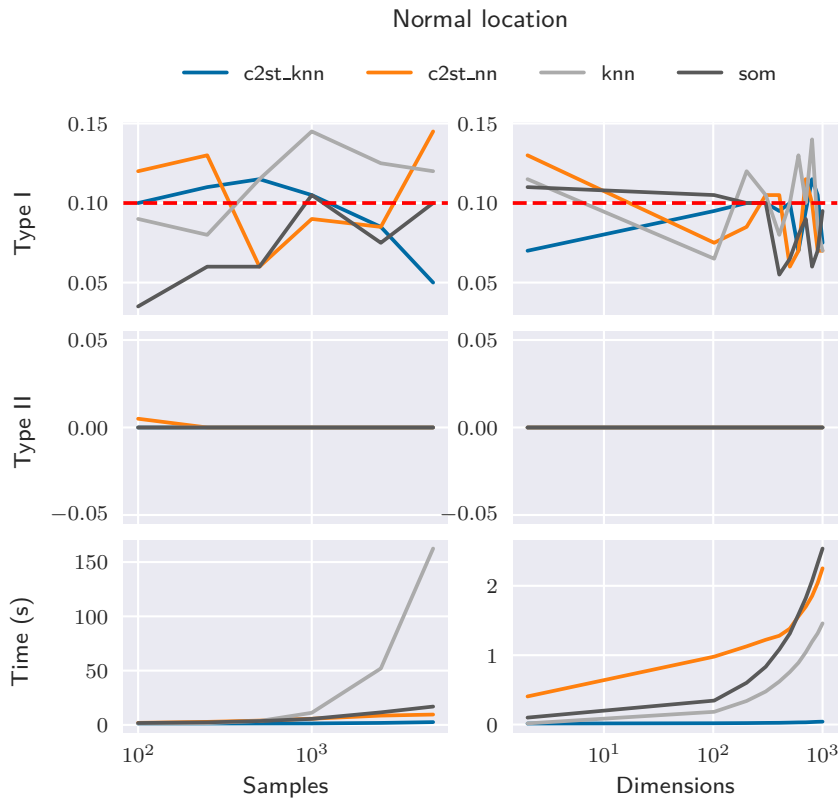}
    \caption{
    Location test for two multivariate Gaussian distributions with $D=1000$}
    \label{fig:normal_location}
\end{figure}

Figure~\ref{fig:normal_location} shows the error rates and run-time for a location
test of two multivariate Gaussian distributions with $D=1000$. For the first distribution,
all dimensions have a mean of $0$, while for the second one, all dimensions have a mean of $0$.

All tests can easily reject $H_0$ within a reasonable run-time. For
a high number of samples, however, the kNN permutation test worsens its run-time performance,
probably due to the imbalance of the KDTree.

\begin{figure}[htbp]
    \centering
    \includegraphics{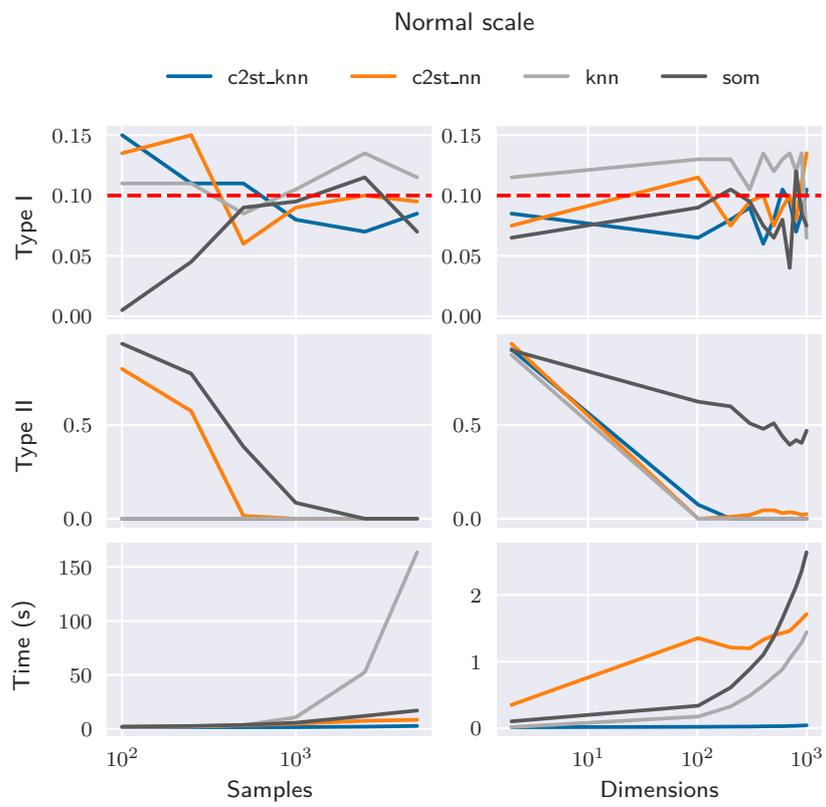}
    \caption{Normal scale}
    \label{fig:normal_scale}
\end{figure}

Figure~\ref{fig:normal_scale} shows the same variables for a scale test of two multivariate
Gaussian distributions with $D=1000$ and two random co-variances matrices samples from the
Wishart distribution~\cite{smith1972algorithm}. In this case, while the type I errors are
well bounded by the significance level, both algorithms based on neural networks
(C2ST-NN and SOM) require a higher number of samples to be able to reject $H_0$, with respect
to the other methods.

Ramdas et al.~\cite{ramdas2015decreasing} have argued that a ``fair'' evaluation of the
power of multivariate non-parametric tests as the dimensionality increases is to keep the
amount of information fixed. i.e. the Kullback-Leibler (KL) divergence between both distributions
should remain constant.

For the location test, this can be achieved by two multivariate Gaussians that only differ
on the first dimension (i.e. $(1, 0, \ldots, 0)$ vs $(0, 0, \ldots, 0)$.

\begin{figure}[htbp]
    \centering
    \includegraphics{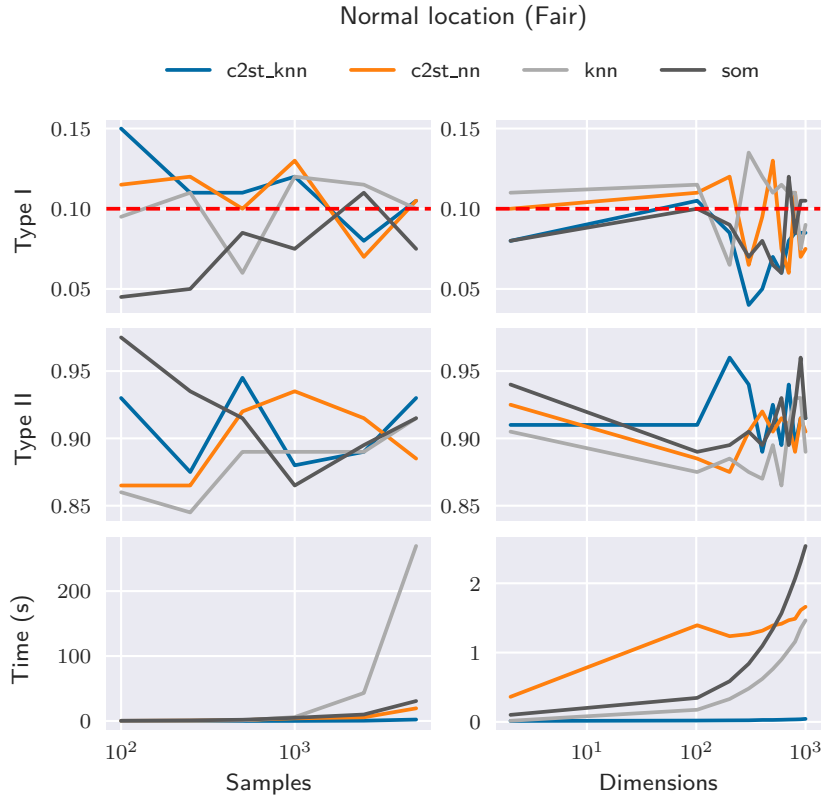}
    \caption{Normal fair location}
    \label{fig:normal_fair_location}
\end{figure}

For completeness, figure~\ref{fig:normal_fair_location} shows the performance of the ML two-sample
tests being evaluated under this condition. We can see that they all fail to improve
their type II error as the dimensionality increase.

\begin{figure}[htbp]
    \centering
    \includegraphics{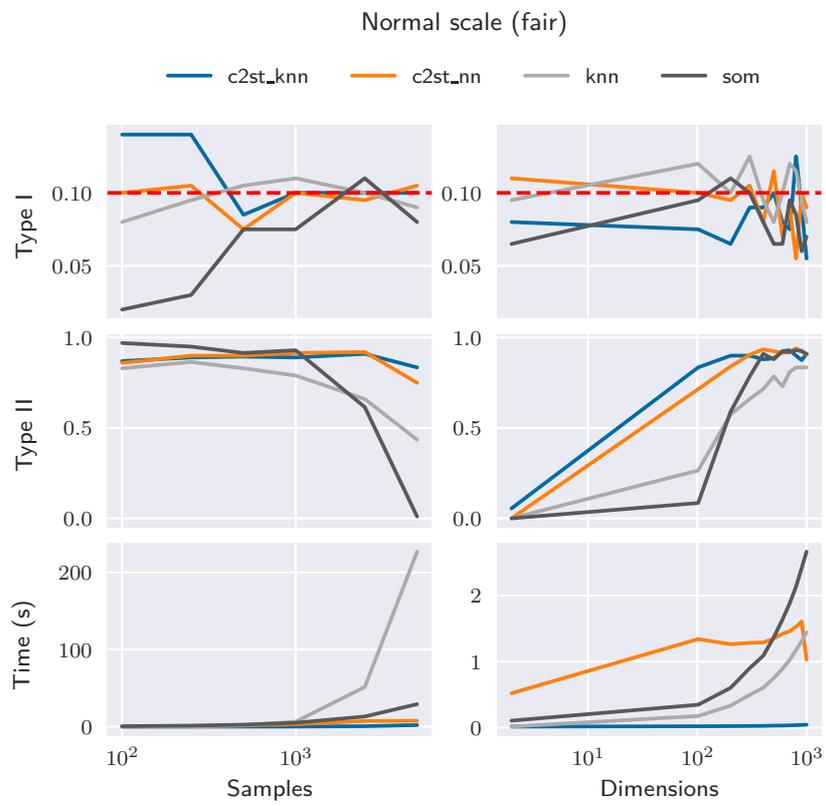}
    \caption{Normal fair scale}
    \label{fig:normal_fair_scale}
\end{figure}

Finally, figure~\ref{fig:normal_fair_scale} shows the performance of the different
tests when only the scale of the first dimension differs, as Ramdas et al. propose
as a fair scale test. In this case, it is more evident that the type II error of all
tests worsens as the number of dimensions increases.

We consider that Ramdas et al. raise a valid point: given the same amount of available
information, additional dimensions do not help. However, for our \PresQ use case, this is an
unrealistic scenario: we are \emph{discovering} the matching dimensions, and each additional
feature will help discriminate whether two samples follow the same distribution.

\subsection{DC2}
The datasets from this challenge come from a single catalog of astronomical
objects split based on the sky coordinates~\cite{EuclidDesprez2020}.

We generate three different samples:

\begin{enumerate}
    \item Samples from the full catalog
    \item Samples applying a magnitude cutout ($\text{MAG}_\text{VIS} < 22.)$
    \item Samples applying a signal-to-noise (SNR) cutout ($\text{VIS} / \text{VIS}_\text{Error} > 10.$)
\end{enumerate}

\begin{figure}[htbp]
    \begin{subfigure}[]{0.5\textwidth}
    \includegraphics[width=\textwidth]{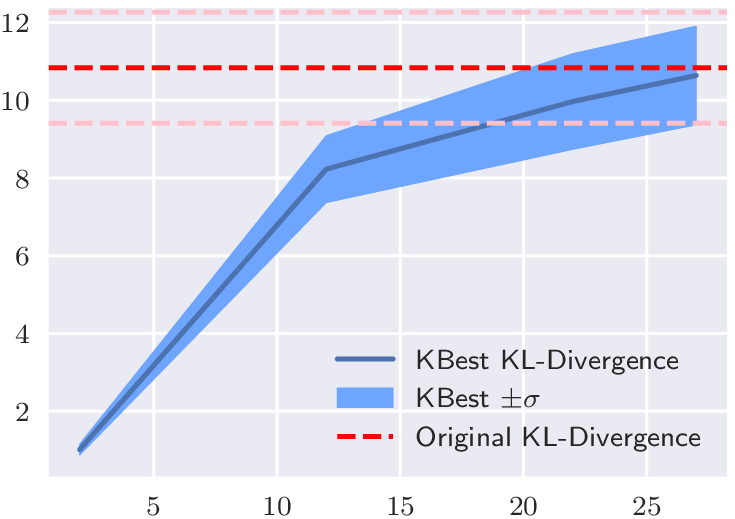}
    \caption{KL Divergence for the $\text{MAG}_\text{VIS}$ cutout}
    \end{subfigure}
    \hfill
    \begin{subfigure}[]{0.5\textwidth}
    \includegraphics[width=\textwidth]{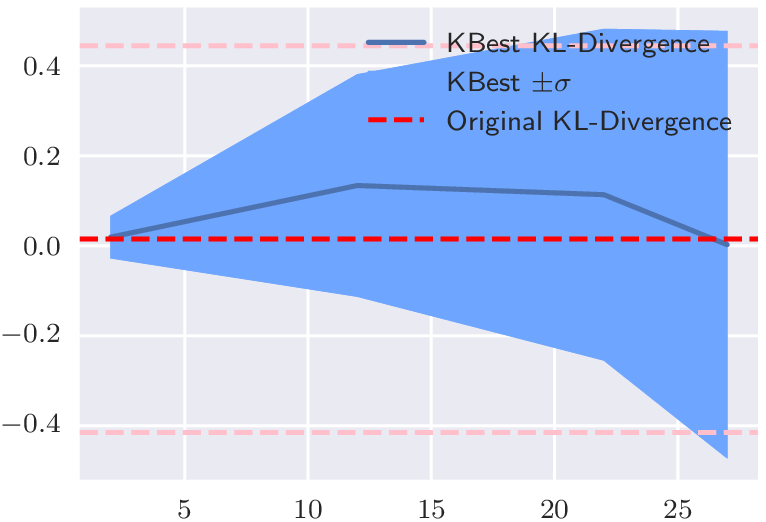}
    \caption{KL Divergence for the SNR cutout}
    \end{subfigure}
    \caption{KL Divergences for the DC2 samples}
    \label{fig:divergence_dc2}
\end{figure}

Following Ramdas et al. paper, in figure~\ref{fig:divergence_dc2}, we report
the estimated KL divergence~\cite{perez2008kullback} between the datasets for an
increasing number of dimensions. It can be seen that the amount of available information
rapidly increases for the magnitude cutout but barely for the SNR one.

\begin{figure}[htpb]
    \centering
    \includegraphics{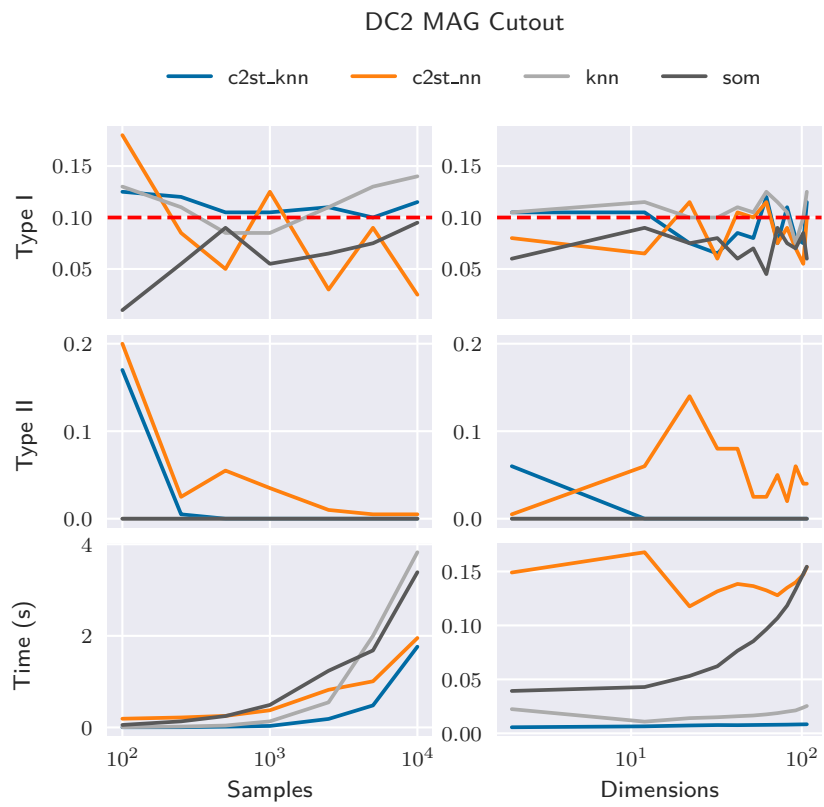}
    \caption{Performance vs sample size (left) and dimensionality (right)}
    \label{fig:dc2_mag}
\end{figure}

Figure~\ref{fig:dc2_mag} shows the measured performances for the DC2 with the magnitude
cutout. Even for a small sample size, the SOM test achieves very low type II errors, 
significantly better than the tests based on classifiers.
This may be because the SOM and kNN
tests can use the full sample, while the classifiers must split the data into training and testing sets.

\begin{figure}[htbp]
    \centering
    \includegraphics{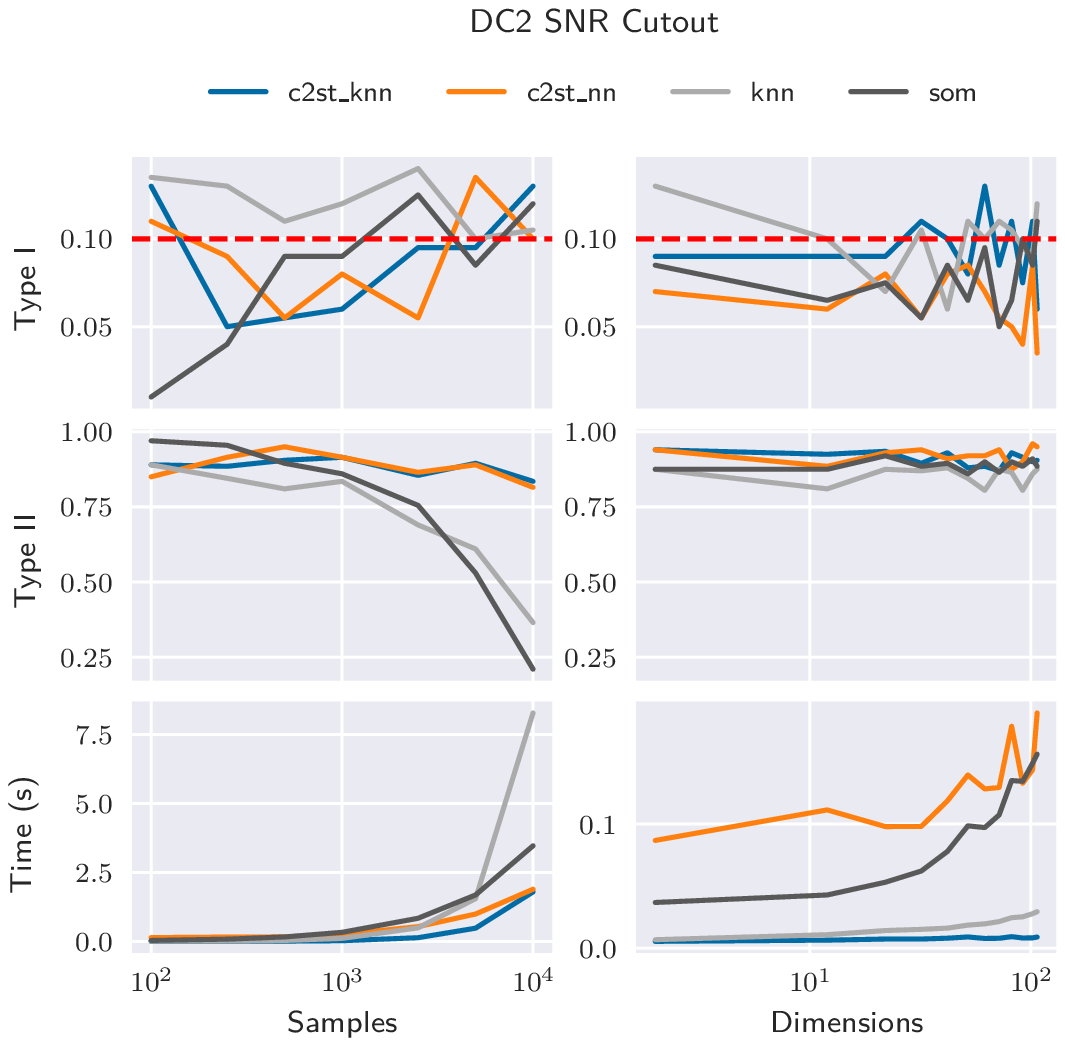}
    \caption{Performance vs sample size (left) and dimensionality (right)}
    \label{fig:dc2_snr}
\end{figure}

Figure~\ref{fig:dc2_snr} shows the same performance metrics for the SNR cutout. In this case,
as suggested by the KL divergence, adding dimensions does not help any of the tests.
However, as the sample size increases, both the kNN and SOM tests achieve a good type II error rate,
while both classifiers remain with a high error rate.

\subsection{Open University Learning Analytics dataset}
\label{subsec:oulad}
The objective of this experiment is to prove that our proposed test can be successfully
used to test a hypothesis, providing an interpretable result useful for further
investigating the data.

We base our test on the \emph{Open University Learning Analytics}
dataset, which contains anonymized data about student demographics~\cite{kuzilek_open_2017}.

Let us consider the case of a researcher with the hypothesis that gender, age, and
region of origin influence a student's economic situation, or perhaps they could
be trying to deanonymize the data.

We can use the \emph{Indices of Multiple Deprivations} (IMD) from this dataset to measure poverty.
The null hypothesis $H_0$ would be that a sample from the overall population and a sample from
the poorest segment are indistinguishable, i.e., they come from the same distribution.

We take all students from the lower end of the poverty line and a sample of the same size
from the general population to test this hypothesis. We run the test using a SOM of size
$20\times20$. The null hypothesis is rejected with a p-value of $0$.

\begin{figure}[t]
    \centering
    \includegraphics[width=\textwidth]{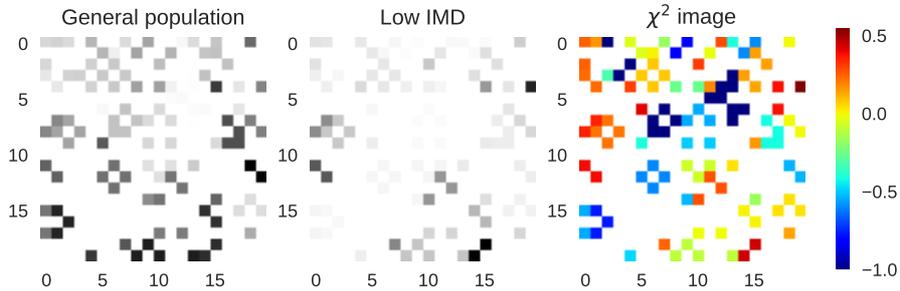}
    \caption{Density of samples for the general population (left), poorest segment (center) and
    relative difference (right). Cells with a value of $-1$ do not have any low-income students,
    while those with a value of $0.5$ show an ``excess'' with respect to the general
    population.}
    \label{fig:oulad_grid}
\end{figure}

Unlike other statistical tests, in addition to the p-value, the researcher can use
the result of the SOM test to compare the projections of both samples.
Figure~\ref{fig:oulad_grid} shows the density of samples for each cell for the overall population
(left), the density of samples from the low-income students (center), and the relative difference
between both (right). The ``most different'' cells hint at how they are different.

\begin{figure}[htbp]
    \centering
    \includegraphics[width=\textwidth]{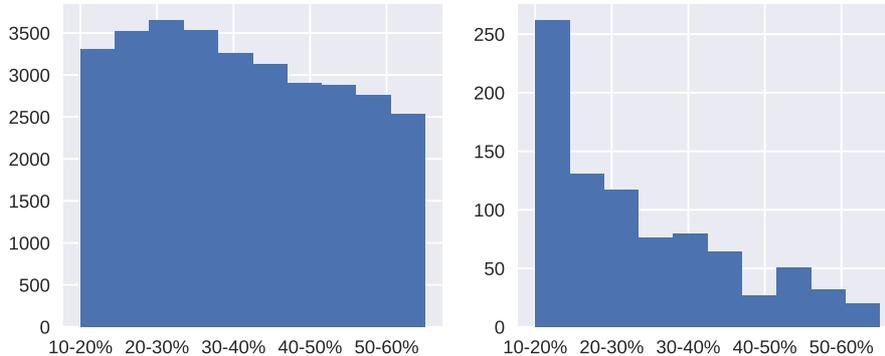}
    \caption{\emph{Indices of Multiple Deprivations} for the general population (left) and for
    young female students from the \emph{North Western Region}.}
    \label{fig:oulad_hist}
\end{figure}

If we pick one of the cells with the most significant bias towards low-income students,
we can see it contains only young female students from the \emph{North Western Region}.
In figure \ref{fig:oulad_hist}, we show the distribution of IMD for the overall population (left),
and for this subset (right). Indeed, the income distribution for this demographics is
heavily skewed towards the low end. We could obtain this hindsight without prior knowledge of which
attributes correlate with the difference, only with the ``hunch'' that
there is a relation. Thus, we prove that our proposed test can be useful for data exploration.

\subsection{Eye Movements}
\label{subsec:eye}
For generating this dataset, 11 subjects were shown a question and a list of ten 
associated sentences, of which one was the correct answer (C), four relevant (R), and five
irrelevant (I). Their eye movement was measured for each possible answer.
The overall measurements were summarized into 22 features, together with the
appropriate label for the sentence~\cite{salojarvi2005inferring}.

With this dataset, we aim to prove that our method can be competitive with other
start-of-the-art proposals, with the benefit of providing a trained model useful for
later purposes.

To evaluate the power of comparing different sets of measures, we replicate Song's
set up and run 2000 times\footnotemark the classifiers and SOM tests using a significance
level of $\alpha = 0.001$ for different sample sizes.
We report their statistical power in table \ref{tab:eye}. We have extracted the kernel-based
results from Song \etal paper~\cite{song2021fast}, including the second-best kernel
based method MMD-B~\cite{zaremba2013b}.

To evaluate the power of comparing different sets of measures, we replicate
Song's setup and run 2000 times the classifiers and SOM tests
using a significance level of $\alpha = 0.001$ for different sample sizes.
We report their statistical power in table \ref{tab:eye}. The results from Song and
MMD-B are extracted from their paper~\cite{song2021fast}.

\footnotetext{Song uses 1000 repetitions.}

\begin{table}[htbp]
    \centering
    \begin{tabular}{c r r r r r r}
        \hline
        \multicolumn{7}{c}{\thead{I vs. C}} \\
        \hline
        \thead{m = n} & \thead{Song} & \thead{MMD-B} & \thead{KNN} & \thead{C2ST-KNN} & \thead{C2ST-NN} & \thead{SOM} \\
        \hline
        100 & 0.826 & 0.374 & \textbf{0.973} & 0.164 & 0.079 & 0.042 \\
        200 & 0.998 & 0.850 & \textbf{1.000} & 0.565 & 0.349 & 0.947 \\
        300 & \textbf{1.000} & 0.985 & \textbf{1.000} & 0.863 & 0.644 & \textbf{1.000} \\
        400 & \textbf{1.000} & \textbf{1.000} & \textbf{1.000} & 0.968 & 0.882 & \textbf{1.000} \\
        \\
        \hline
        \multicolumn{7}{c}{\thead{R vs. C}} \\
        \hline
        \thead{m = n} & \thead{Song} & \thead{MMD-B} & \thead{KNN} & \thead{C2ST-KNN} & \thead{C2ST-NN} & \thead{SOM} \\
        \hline
        100 & 0.670 & 0.236 & \textbf{0.845} & 0.062 & 0.023 & 0.007 \\
        200 & 0.969 & 0.685 & \textbf{0.996} & 0.298 & 0.139 & 0.672 \\
        300 & 0.999 & 0.941 & \textbf{1.000} & 0.558 & 0.314 & 0.987 \\
        400 & \textbf{1.000} & 0.988 & \textbf{1.000} & 0.811 & 0.560 & \textbf{1.000} \\
    \end{tabular}
    \caption{Empirical statistical power for the eye movement datasets}
    \label{tab:eye}
\end{table}

The results show that our test has low power for small samples, but it rapidly gains
terrain compared to the classifier-based methods, being competitive even with the
kernel-based methods. The nearest-neighbors method is remarkably efficient in all cases.

To evaluate the usefulness of the trained model obtained as part of running the
statistical test, we use the trained SOM as a classifier by simply labeling the input
with a majority rule applied to each SOM cell. We performed a 50-fold cross-validation
with a sample size of $n=m=409$, so each training set is $n=m\approx 400$, corresponding
to the last entries in table \ref{tab:eye}.

The obtained mean accuracy where: C vs. I 68.42\%; C vs. R 67.84\%; I vs R 53.90\%.
These results, even with a relatively small sample size, are on par with those reported
on the paper from which the dataset was obtained~\cite{salojarvi2005inferring}.

\begin{figure}[htb]
    \centering
    \includegraphics[width=\textwidth]{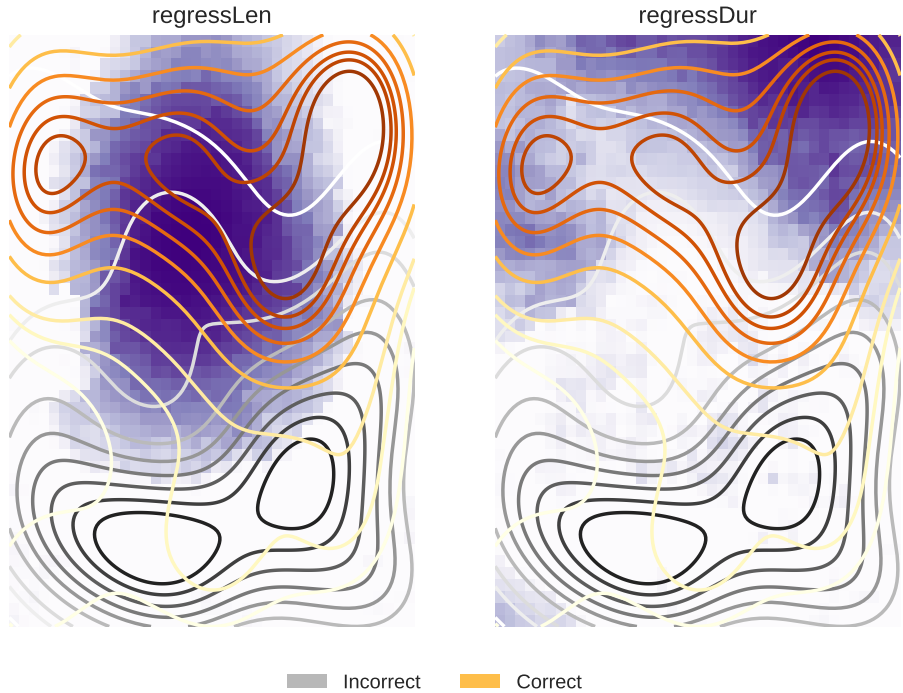}
    \caption{Codebook values for the two most distinct features from the Eye dataset.
    Darker neurons are more sensitive to the given feature.
    The lines represent the density estimation for the Correct and \emph{Incorrect} samples over the SOM.
    It is visible that the Incorrect category ``wraps'' around long regression distances, while the
    \emph{Correct} category correlates positively with the regression duration.}
    \label{fig:eye_distinct_features}
\end{figure}

Finally, as an exercise on interpretability, figure~\ref{fig:eye_distinct_features} shows
the value of the two most distinct codebook dimensions. These
attributes, related to the regression (re-reading a word), show a sharp distinction
that matches the distribution of samples from the Correct and Incorrect samples quite well.
Indeed, this matches the expectations from the original paper that the second-pass
measures indicate high-level cognitive processing and correlate with
conscious efforts when choosing a correct answer.

\subsection{Age (IMDb Faces)}
The IMDb-WIKI dataset~\cite{rothe2018deep} contains 460,723 face images
extracted from IMDb. The authors additionally provide a neural network pre-trained
to predict the age of the face cutouts. Similarly to Song's experiments~\cite{song2021fast},
we have run the neural model over the IMDb faces, extracting the values from
the last hidden layer (4096 neurons) as the target multivariate distribution.

We then group the samples in age ranges and verify our proposal, comparing
500 samples from consecutive age groups, and repeat 500 times for each experiment. The
significance level is also set to $0.001$.

Table \ref{tab:age} shows the results of the SOM test, together
with the results reported by Song\etal for their kernel-based method and for
MMD-B\cite{zaremba2013b}.

\begin{table}[htbp]
    \centering
    \begin{tabular}{lrrrr}
    \hline
    Age ranges & \thead{Song} & \thead{MMD-B} & \thead{KNN} & \thead{SOM} \\
    \hline
    15-20 vs. 20-25 & 1.000 & 1.000 & 1.000 & 1.000 \\
    20-25 vs. 25-30 & 1.000 & 0.800 & 0.984 & 0.990 \\
    25-30 vs. 30-35 & 0.990 & 0.790 & 0.876 & 0.726 \\
    30-35 vs. 35-40 & 1.000 & 0.830 & 0.866 & 0.812 \\
    35-40 vs. 40-45 & 0.950 & 0.250 & 0.784 & 0.564 \\
    40-45 vs. 45-50 & 0.930 & 0.400 & 0.812 & 0.606 \\
    \hline
    \end{tabular}
    \caption{Age}
    \label{tab:age}
\end{table}

For this particular experiment Song's proposal outperforms both MMD-B and the SOM
test, although our method comes second in statistical power.


\subsection{\PresQ}
As we have mentioned in the introduction, the motivating example is the unsupervised
discovery of shared attributes between multiple datasets using \PresQ \cite{alvarez2021noise},
obtaining, as a result, the list of sets of attributes and a trained model that can be used to
cross-match the datasets.

We have run two examples from the \PresQ paper using the proposed SOM-based test
instead of the $k$NN statistical test. For each of the examples, we measure the following:

\begin{description}
    \item[Ratio] of known shared attributes identified by \PresQ
    \item[Overhead] Number of tests per unique Equally-Distributed Dependency found
    \item[Time] that took \PresQ to finish, including the time taken for serializing the SOM models
\end{description}

\begin{figure}[ht]
    \centering
    \includegraphics[width=\textwidth]{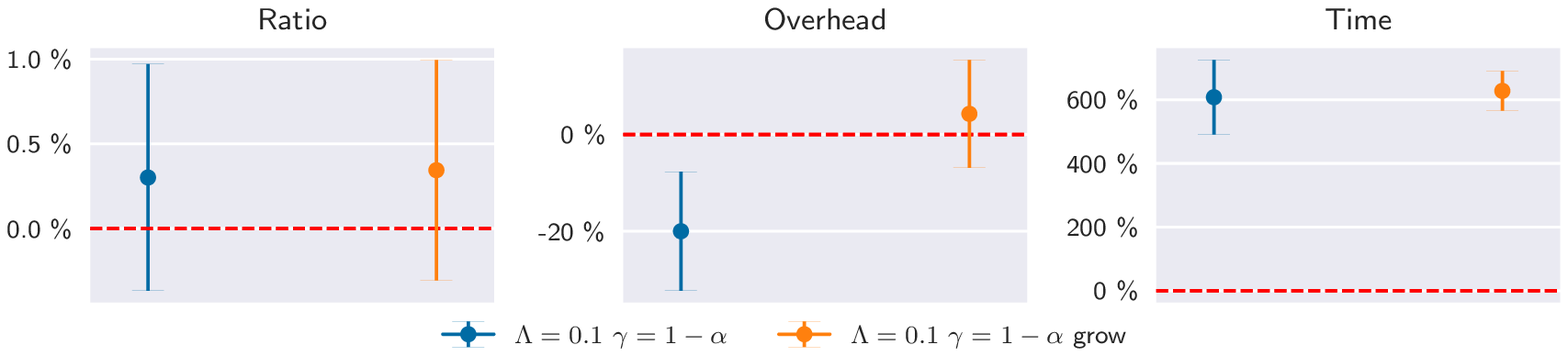}
    
    \includegraphics[width=\textwidth]{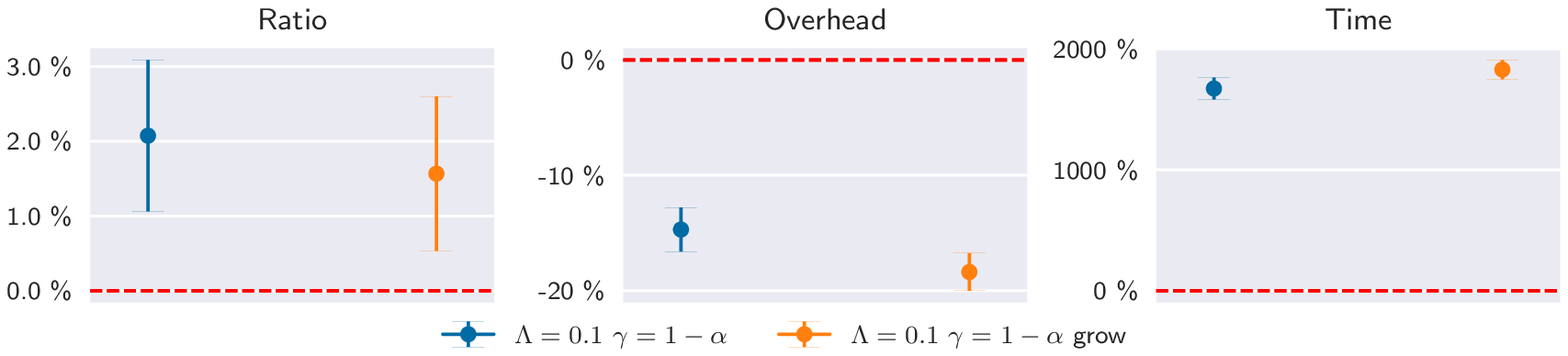}
    \caption{Relative difference between $\PresQ_{SOM}$ and $\PresQ_{kNN}$.
    Top: DC2 dataset~\cite{EuclidDesprez2020}.
    Bottom: Ailerons / Elevators datasets~\cite{alcala2011keel}.}
    \label{fig:presq_som}
\end{figure}

Figure~\ref{fig:presq_som} reports the 95\% confidence interval ($\mu \pm 1.96 \sigma$)
for the \emph{relative differences} obtained when running with SOM vs
$k$NN. Their distribution has been estimated using bootstrapping. The parametrization for
\PresQ was $\Lambda = 0.1$, $\gamma=0.95$, $\alpha=0.05$ and a sample size of 1000.

The SOM test has a run-time penalty because it is a more complex model to train.
However, fewer tests are required for the same number of unique EDD found. This is likely a consequence of
the $kNN$ test being slightly more prone to reject the equality of distribution than the SOM
test, as we can see in figures \ref{fig:normal_location}, \ref{fig:normal_scale}.

\section{Discussion}
\label{sec:discussion}
The results shown in section~\ref{sec:results} prove that the non-parametric, two-sample SOM
test described in section~\ref{sec:chi2} generally outperforms, in terms of power, other classifier-based
two sample tests, being in some cases even comparable to kernel-based methods. It has the added
advantage of generating an interpretable and usable model: for instance, the resulting SOM
could be used as the basis for a SOM-kNN combined classification model~\cite{silva2011som}.

Like other machine-learning or kernel-based approaches, our method requires some initial parameters,
such as the SOM size, to be set by the user. From our tests, networks of size $O(100)$ neurons
generally work well enough, but for more precise control, the SOM size can be set to the lengths
of the two largest principal components~\cite{KOHONEN201352}.

It has been argued that provable, \emph{proper}, test statistics should be, in general,
preferred over heuristic alternatives~\cite{rosenblatt2021better}. However, our proposal is
more oriented toward exploring abundant structured data. In this case, developing a
tailored statistic for all possible combinations is not viable, and a
pragmatic approach is more suitable~\cite{kim2021classification}.

Finally, if the null hypothesis - that both samples come from the same underlying distribution -,
is rejected, the generated SOM model can be easily visualized and examined in more detail.
In sections~\ref{subsec:oulad} and \ref{subsec:eye} we have demonstrated that we can
reject $H_0$ and obtain valuable
information after exploring the learned model. This exercise would not have
been possible with a black-box method as the neural network classifier~\cite{friedman2004multivariate}.

As a side note, SOM maps can be generalizable to non-vectorial data --- i.e., strings ---
as long as more than one ordering relation is defined~\cite{kohonen1982self, KOHONEN201352}.
Therefore, our proposed statistical test could also be used to test whether two sets of sequences
--- i.e., proteins or DNA --- share their origin. Unfortunately, since the SOM implementation we
have based our implementation on~\cite{wittek2013somoclu} only works with real-valued dimensions,
we have not been able to test this scenario.

\section{Conclusion and future work}
\label{sec:conclusions}
As part of interactive data exploration, researchers may need to compare multiple datasets.
These datasets can originate from various independent files or generative models that need
to be compared with reality. When these datasets are of high dimensionality, especially if
the exploration is tentative, developing tailored statistical tests can become impractical.
In those cases, relying on heuristic approaches based on machine learning techniques as
classifiers to decide whether two samples are distinguishable becomes a good
alternative~\cite{friedman2004multivariate,kim2021classification}.

However, some of these methods, like neural networks, are hard to interpret when rejecting
the null hypothesis $H_0: P = Q$. In other words, they reject that both samples originate
from the same underlying distribution but do not assist the researcher any further.
Other models, for instance, random forests, are more  interpretable~\cite{friedman2004multivariate}.

In section~\ref{sec:chi2}, we have proposed another machine-learning technique based on
Self-Organizing Maps~\cite{kohonen1982self} that is understandable and capable of pointing
the researcher to where the differences on a multidimensional regime are. After all, SOMs
were initially proposed as visualization aids. Nonetheless, they display interesting emergent
properties and can be used for clustering or classification as well~\cite{ultsch2007emergence}.

In section~\ref{sec:results}, we have proven that the power of this technique is comparable
to other machine learning techniques and even superior for medium-size datasets.
We have also proven in the experiment~\ref{subsec:oulad} that the output can guide
researchers toward refining a hypothesis. Thus, our test can be a valuable asset to
the researcher's tool-set, and complementary to more formal hypothesis testing whenever
considered necessary~\cite{rosenblatt2021better}.

For future work, it would be interesting to explore the possibilities of properties of emergent
Self-Organizing Maps to assist the researcher in examining the differences.
For instance, clustering could be helpful to identify whole regions that differ rather than
focusing on individual neurons.

\FloatBarrier

\bibliographystyle{ieeetr}
\bibliography{references.bib}

\newpage
\appendix

\section{Additional plots}
\label{app:plots}

\subsection{KL-Divergences}

\begin{figure}[hp]
    \begin{subfigure}[]{0.5\textwidth}
    \includegraphics[width=\textwidth]{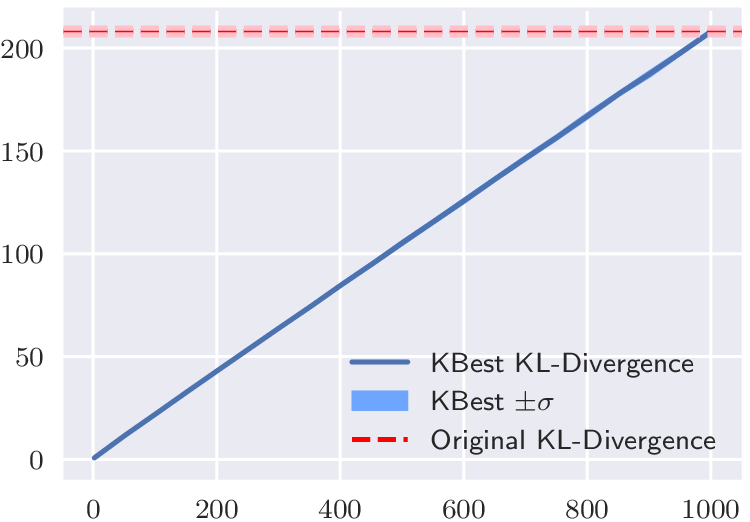}
    \caption{Normal}
    \end{subfigure}
    \hfill
    \begin{subfigure}[]{0.5\textwidth}
    \includegraphics[width=\textwidth]{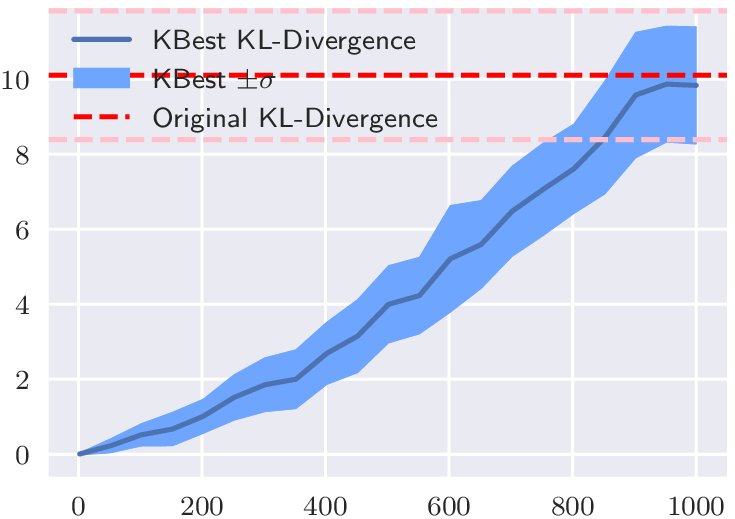}
    \caption{Scale}
    \end{subfigure}
    \label{fig:divergence_normal}
    \caption{KL Divergence for a normal distribution}
\end{figure}

\begin{figure}[htbp]
    \begin{subfigure}[]{0.5\textwidth}
    \includegraphics[width=\textwidth]{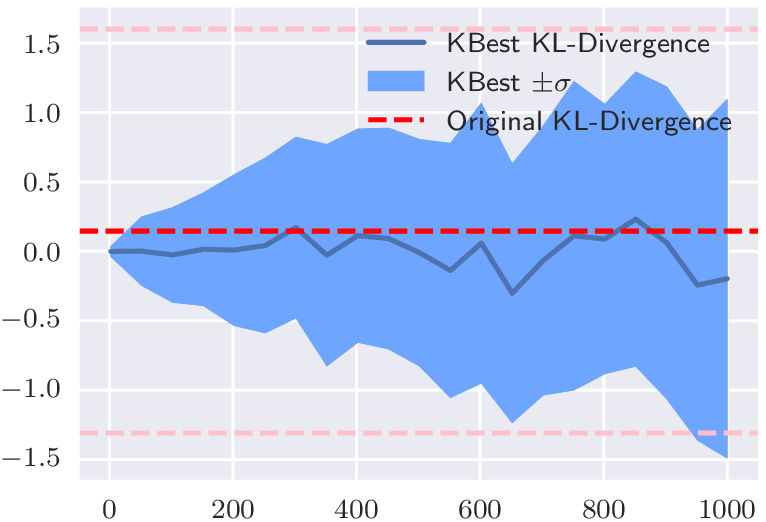}
    \caption{Normal}
    \end{subfigure}
    \hfill
    \begin{subfigure}[]{0.5\textwidth}
    \includegraphics[width=\textwidth]{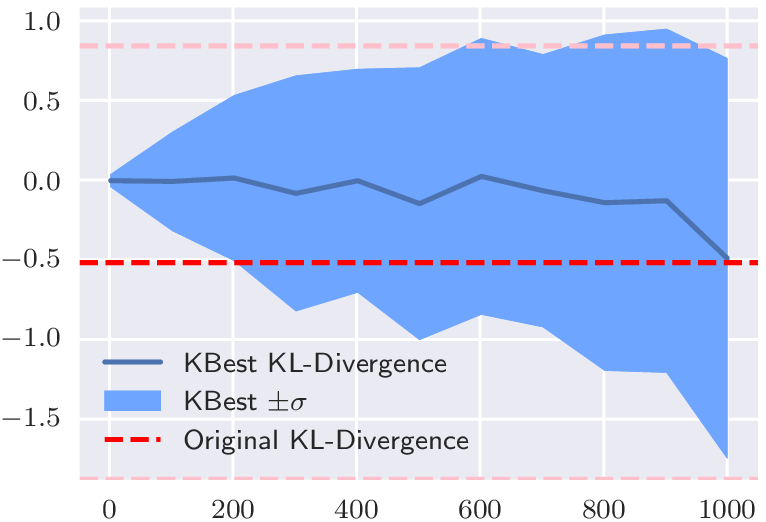}
    \caption{Scale}
    \end{subfigure}
    \label{fig:divergence_fair}
    \caption{KL Divergence for a normal ``fair'' distribution}
\end{figure}

\FloatBarrier
\newpage
\subsection{MNIST}

\begin{figure}[htbp]
    \centering
    \includegraphics[width=0.5\textwidth]{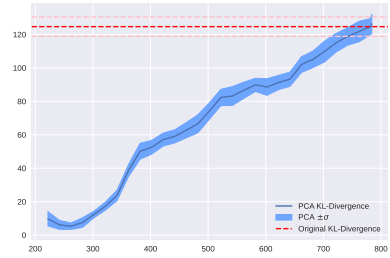}
    \label{fig:divergence_mnist}
    \caption{KL Divergence for MNIST}
\end{figure}

\begin{figure}[htbp]
    \centering
    \includegraphics[width=0.8\linewidth]{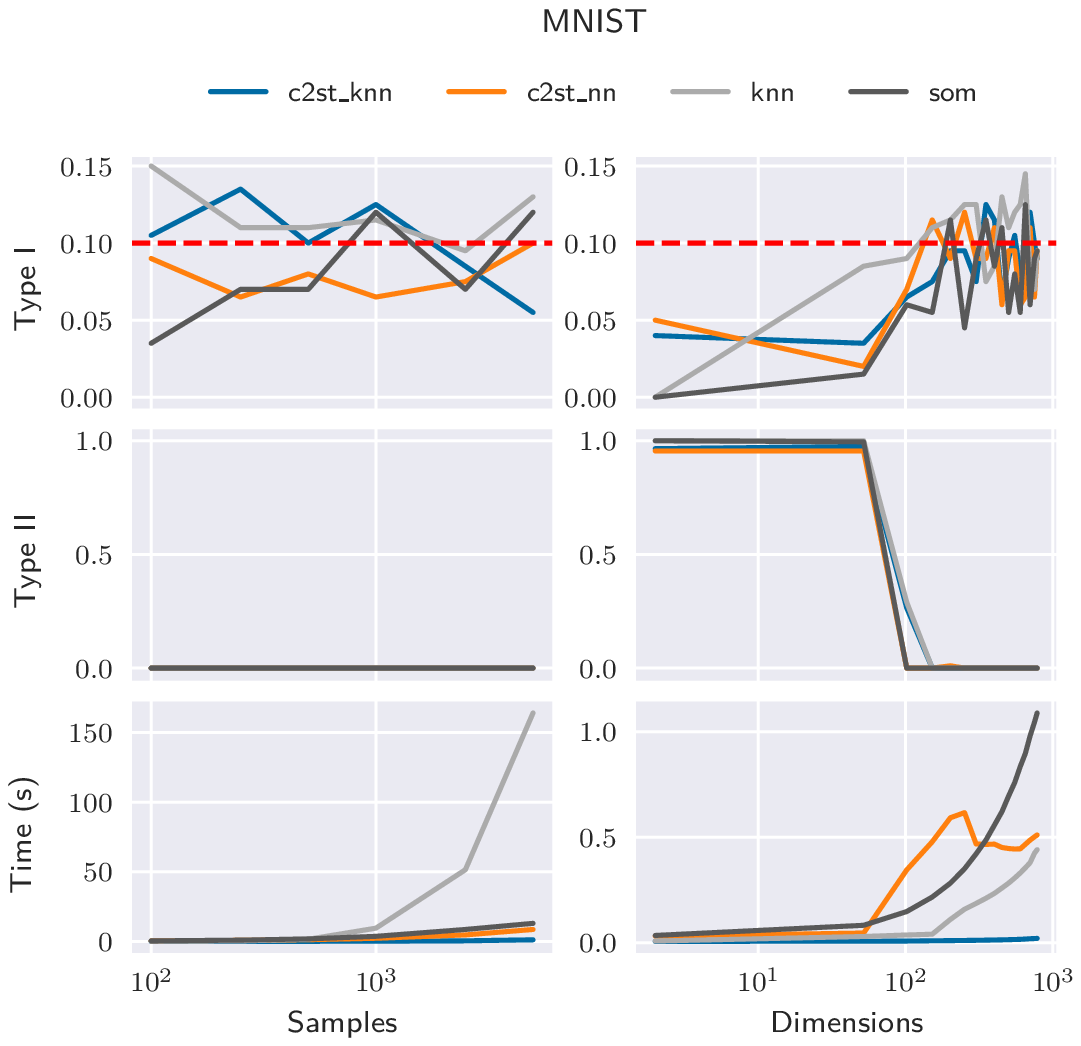}
    \caption{MNIST}
    \label{fig:mnist}
\end{figure}

\FloatBarrier
\newpage

\subsection{Eye plots}

\begin{figure}[htpb]
    \centering
    \includegraphics[width=0.8\linewidth]{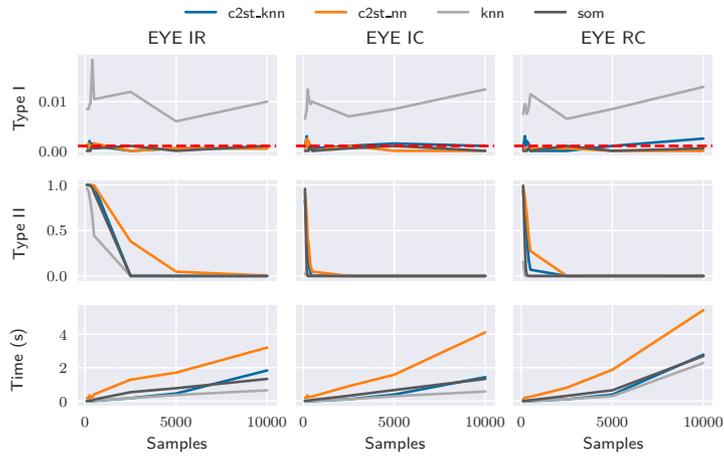}
    \caption{Eye dataset}
    \label{fig:sample_eye}
\end{figure}

\begin{figure}[htpb]
    \centering
    \includegraphics[width=0.85\linewidth]{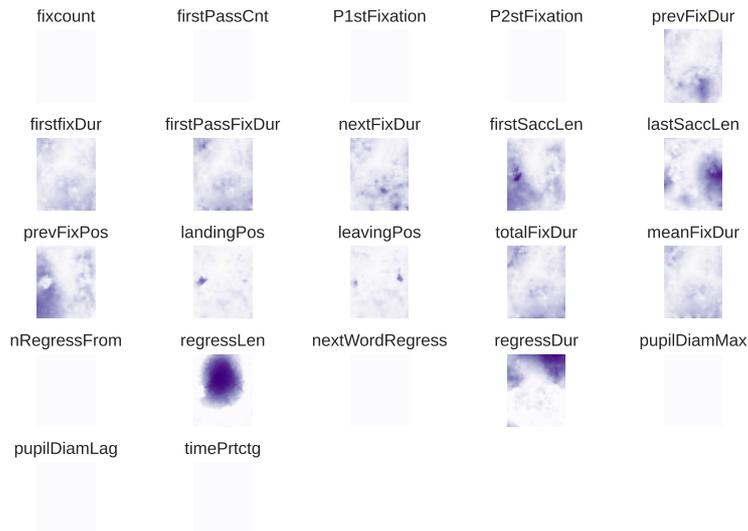}
    \caption{Codebook values for the different features from the Eye dataset}
    \label{fig:eye_features}
\end{figure}

\end{document}